\documentclass[11pt,a4paper]{article}
\usepackage[hyperref]{emnlp2018}
\usepackage{times}
\usepackage{latexsym}
\usepackage{booktabs}
\usepackage{graphicx}
\graphicspath{ {images/} }

\usepackage{url}

\aclfinalcopy 

\title{
How Much \emph{Reading} Does Reading Comprehension Require? \\ 
A Critical Investigation of Popular Benchmarks
}

\author{Divyansh Kaushik \\
 Language Technologies Institute\\
Carnegie Mellon University\\
 {\tt dkaushik@cs.cmu.edu} \\\And
 Zachary C. Lipton \\
Tepper School of Business \\
Carnegie Mellon University \\
 {\tt zlipton@cmu.edu} \\}

\date{}

\begin{document}
\maketitle
\begin{abstract}
Many recent papers address \emph{reading comprehension},
where examples consist of (\emph{question}, \emph{passage}, \emph{answer}) tuples. 
Presumably, a model must combine 
information from \emph{both} questions and passages 
to predict corresponding answers.
However, despite intense interest in the topic,
with hundreds of published papers vying for leaderboard dominance, 
basic questions about the difficulty 
of many popular benchmarks remain unanswered.
In this paper, we establish sensible baselines
for the bAbI, SQuAD, CBT, CNN, 
and Who-did-What datasets,
finding that question- and passage-only
models often perform surprisingly well.
On $14$ out of $20$ bAbI tasks,
passage-only models 
achieve greater than $50\%$ accuracy,
sometimes matching the full model.
Interestingly, while CBT provides $20$-sentence stories
only the last is needed for comparably accurate prediction.
By comparison, SQuAD and CNN appear better-constructed.
\end{abstract}


\section{Introduction}
\label{sec:intro}
Recently, \emph{reading comprehension} (RC) 
has emerged as a popular task, 
with researchers proposing 
various end-to-end deep learning algorithms 
to push the needle on a variety of benchmarks.
As characterized by \citet{hermann2015teaching,onishi2016did},
unlike prior work addressing 
question answering from general structured knowledge,
RC requires that a model extract information 
from a given, unstructured passage. 
It's not hard to imagine how such systems could be useful.
In contrast to generic text summarization,
RC systems could answer targeted questions about specific documents, 
efficiently extracting facts and insights.

While many RC datasets have been proposed 
over the years \citep{hirschman1999deep,breck2001looking,penasoverview11,penasoverview12,sutcliffeoverview,richardson2013mctest,berant2014modeling}, 
more recently, larger datasets have been proposed 
to accommodate the data-intensiveness of deep learning.
These vary both in the source and size of their corpora 
and in how they cast the prediction problem---as a classification task \citep{hill2015goldilocks,hermann2015teaching,onishi2016did,lai2017race,babi,miller2016key}, 
span selection \citep{rajpurkar2016squad,trischler2017newsqa}, 
sentence retrieval \citep{wang2007jeopardy,yang2015wikiqa},
or free-form answer generation \cite{nguyen2016ms}.\footnote{
We note several other QA datasets 
\citep{yang2015wikiqa,miller2016key,
nguyen2016ms,paperno2016lambada,clark2016my,
lai2017race,trischler2017newsqa,joshi2017triviaqa} 
not addressed in this paper.
}
Researchers have steadily 
advanced on these benchmarks,
proposing myriad neural network architectures 
aimed at attending to both questions and passages 
to produce answers.

In this paper, we argue 
that amid this rapid progress on empirical benchmarks,
crucial steps are sometimes skipped.
In particular, we demonstrate 
that the level of difficulty 
for several of these tasks 
is poorly characterized.  
For example, for many RC datasets, it's not reported, 
either in the papers introducing the datasets, 
or in those proposing models, 
how well one can perform while ignoring either the question or the passage.
In other datasets, although the passage might consist of many lines of text, it's not clear how many are actually required to answer the question, e.g., 
the answer may always lie in the first or the last sentence.

%
%
%


We describe several popular RC datasets and models proposed for these tasks, 
analyzing their performance 
when provided with question-only (Q-only) or passage-only (P-only) information. 
We show that on many tasks,
the results obtained are surprisingly strong,
outperforming many baselines,
and sometimes even surpassing the same models,
supplied with both questions \emph{and} passages.

We note that similar problems 
were shown for datasets in 
\emph{visual question answering} by \citet{goyal2017making} 
and for \emph{natural language inference} by \citet{N18-2017,
poliak_nli, glockner_acl18}.
Several other papers have discussed the weaknesses of various RC benchamrks \citep{chen2016thorough,lee2015reasoning}.
We discuss these studies in the paragraphs introducing the corresponding datasets below.

\section{Datasets}
\label{sec:datasets}
In the following section, 
we provide context on each dataset
that we investigate and then describe our process 
for corrupting the data 
as required by our question- and passage-only experiments. 

\paragraph{CBT} 
\citet{hill2015goldilocks} prepared
a cloze-style (\emph{fill in the blank})
RC dataset by using passages from children's books. 
In their dataset, 
each passage consists of $20$ consecutive sentences, 
and each question is
the $21$st sentence 
with one word removed.
The missing word then serves as the answer. 
The dataset is split into four categories of answers: 
Named Entities (NE), Common Nouns (CN), 
Verbs (V) and Prepositions (P). 
The training corpus contains over $37,000$ 
candidates and each question 
is associated with $10$ candidates, 
POS-matched to the correct answer. 
The authors established LSTM/embedding-based Q-only  baselines 
but did not present the results obtained 
by their best model using Q-only or P-only information.

\paragraph{CNN}
\citet{hermann2015teaching}
introduced the CNN/Daily Mail datasets 
containing more than $1$ million news articles,
each associated with several highlight sentences.
Also adopting the cloze-style 
dataset preparation, 
they remove an entity (answer) 
from a highlight (question).
They anonymize all entities to ensure 
that models rely on information 
contained in the passage, 
vs memorizing characteristics 
of given entities across examples, and thus ignoring passages.
On average, passages contain $26$ entities,
with over $500$ total possible answer candidates.
\citet{chen2016thorough} analyzed 
the difficulty of the CNN and Daily Mail tasks. 
They hand-engineered a set of eight features for each entity $e$ (does $e$ occur in the question, in the passage, etc.), showing that this simple classifier outperformed many earlier deep learning results.

\paragraph{Who-did-What}
\citet{onishi2016did} extracted pairs of news articles, each pair 
referring to the same events. 
Adopting the cloze-style, 
they remove a person's name (the answer) from the first sentence of one article (the question).
A model must predict the answer based on the question,
 together with the other article in the pair (passage). 
Unlike CNN, Who-did-What does not anonymize entities. 
On average, each question is associated with $3.5$ candidate answers. 
The authors removed several questions 
from their dataset 
to thwart simple strategies 
such as always predicting the name 
that occurs most (or first) in the passage.

\paragraph{bAbI}
\citet{babi}
presented a set of $20$ tasks 
to help researchers identify and rectify the failings of 
their reading comprehension systems. 
Unlike the datasets discussed so far, the questions in this task are not cloze-style and are synthetically generated using templates. This restricts the diversity in clauses appearing in the passages. Further, this also restricts the dataset vocabulary to just $150$ words, in contrast, CNN dataset has a vocabulary made of close to $120,000$ words.
Memory Networks with adaptive memory, n-grams and non-linear matching were shown to obtain $100\%$ accuracy on $12$ out of $20$ bAbI tasks. 
We note that \citet{lee2015reasoning} 
previously identified that bAbI tasks might fall short as a measure of ``AI-complete question answering'',
proposing two models based on \emph{tensor product representations}
that achieve $100\%$ accuracy on many bAbI tasks.

\paragraph{SQuAD}
More recently, \citet{rajpurkar2016squad} 
released the Stanford Question Answering Dataset (SQuAD)
containing over $100,000$ crowd-sourced questions 
addressing $536$ passages. 
Each question is associated with a paragraph (passage)
extracted from an article. 
These passages are shorter than those in CNN and Who-did-What datasets.
Models choose answers by selecting (varying-length) 
spans from these passages.

\paragraph{Generating Corrupt Data}
To void any information in either the questions or the passages, while otherwise leaving each architecture intact,
we create corrupted versions of each dataset by assigning either
questions randomly, while preserving the correspondence between passage and answer, or by randomizing the passage.
For tasks where question-answering requires selecting spans or candidates from the passage, we create passages that contain the candidates in random locations but otherwise consist of random gibberish.


\section{Models}
\label{sec:models}
In our investigations of the various RC benchmarks, 
we rely upon the following three recently-proposed models: key-value memory networks, gated attention readers, and QA nets. Although space constraints preclude a full discussion of each architecture, we provide references to the source papers and briefly discuss any implementation decisions necessary to reproduce our results.

\paragraph{Key-Value Memory Networks} 
We implement a Key-Value Memory Network (KV-MemNet) \citep{miller2016key}, 
applying it to bAbI and CBT. 
KV-MemNets are based on Memory Networks \citep{sukhbaatar2015end},
shown to perform well on both datasets. 
For bAbI tasks, the keys and values 
both encode the passage as a bag-of-words (BoW). 
For CBT, the key is a BoW-encoded $5$-word window 
surrounding a candidate answer  
and the value is the candidate itself. 
We fixed the number of hops to $3$ 
and the 
embedding size to $128$.

\paragraph{Gated Attention Reader} 
Introduced by \citet{dhingra2017gated},
the Gated Attention Reader (GAR)\footnote{https://github.com/bdhingra/ga-reader} performs multiple hops over a passage,
like MemNets. 
The word representations are refined over each hop 
and are mapped by an attention-sum module \cite{kadlec2016text} 
to a probability distribution 
over the candidate answer set in the last hop. 
The model nearly matches best-reported results 
on many cloze-style RC datasets, 
and thus we apply it to \emph{Who-did-What, CNN, CBT-NE} and \emph{CBT-CN}.

\paragraph{QA Net}
Recently introduced by \cite{Yu2018QANetCL},
the QA-Net\footnote{We use the implementation available at https://github.com/NLPLearn/QANet} 
was recently demonstrated to outperform 
all previous models on the SQuAD dataset\footnote{At 
the time of publication, 
an ensemble of QA-Net models 
was at the top of the leader board. 
A single QA-Net was ranked $4$th.}. 
Passages and questions are passed as input 
to separate encoders consisting of depth-wise separable convolutions 
and global self-attention. 
This is followed by a passage-question attention layer, 
followed by stacked encoders. 
The outputs from these encoders are used 
to predict an answer span inside the passage.

\begin{table*}[t!]
\begin{center}
\renewcommand{\arraystretch}{1.2}
  \begin{tabular}{l lllll lllll}
& \multicolumn{10}{c}{\textbf{bAbI Tasks 1-10}}\\
\cline{2-11}
Dataset & 1 & 2 & 3 & 4&5&6&7&8&9&10\\
\hline
True dataset & $\textbf{100\%}$&$\textbf{100\%}$&$39\%$&$\textbf{100\%}$&$\textbf{99\%}$&$\textbf{100\%}$&$\textbf{94\%}$&$\textbf{97\%}$&$\textbf{99\%}$&$\textbf{98\%}$\\
Question only & $18\%$&$17\%$&$22\%$&$22\%$&$34\%$&$50\%$&$48\%$&$34\%$&$64\%$&$44\%$\\
Passage only & $53\%$&$86\%$&$\textbf{60\%}$&$59\%$&$31\%$&$48\%$&$85\%$&$79\%$&$63\%$&$47\%$\\
$\Delta (min)$ & $-47$&$-14$&$+21$&$-41$&$-65$&$-52$&$-9$&$-18$&$-35$&$-51$\\
%
\midrule
& \multicolumn{10}{c}{\textbf{bAbI Tasks 11-20}}\\
\cline{2-11}
 & 11 & 12 & 13 & 14&15&16&17&18&19&20\\
\midrule
True dataset & $\textbf{94\%}$&$\textbf{100\%}$&$\textbf{94\%}$&$\textbf{96\%}$&$\textbf{100\%}$&$\textbf{48\%}$&$\textbf{57\%}$&$\textbf{93\%}$&$\textbf{30\%}$&$\textbf{100\%}$\\

Question only & $17\%$&$15\%$&$18\%$&$18\%$&$34\%$&$26\%$&$48\%$&$91\%$&$10\%$&$70\%$\\

Passage only & $71\%$&$74\%$&$\textbf{94\%}$&$50\%$&$64\%$&$\textbf{47\%}$&$48\%$&$53\%$&$21\%$&$\textbf{100\%}$\\

$\Delta (min)$ & $-23$&$-26$&$0$&$-46$&$-36$&$-1$&$-9$&$-2$&$-9$&$0$\\
\bottomrule
\end{tabular}
\caption{Accuracy on bAbI tasks using our implementation of the Key-Value Memory Networks}\label{tab:tab_babi}
\end{center}
\end{table*}

\begin{table}[t!]
  \begin{center}
  \renewcommand{\arraystretch}{1.2}
  \begin{tabular}{ l c c c c }
    \midrule
    Task & Full & Q-only & P-only & $\Delta (min)$ \\
    \midrule
    \multicolumn{5}{c}{Key-Value Memory Networks}\\
    \midrule
    CBT-NE & $\textbf{35.0\%}$&	$29.1\%$&	$24.1\%$&$-5.9$\\

    CBT-CN & $\textbf{37.6\%}$&	$32.4\%$&	$24.4\%$&$-5.2$ \\

    CBT-V &$52.5\%$ &	$\textbf{55.7\%}$&	$36.0\%$ &$+3.2$ \\

    CBT-P& $55.2\%$&	$\textbf{56.9\%}$&	$30.1\%$ &$+1.7$\\
	\midrule
    \multicolumn{5}{c}{Gated Attention Reader}\\
    \midrule  
	CBT-NE&	$\textbf{74.9\%}$ & $50.6\%$&	$40.8\%$ & $-17.5$\\
	CBT-CN&	$\textbf{70.7\%}$ &	$54.0\%$&	$36.7\%$&$-16.7$\\
    CNN&$\textbf{77.8\%}$ &	$25.6\%$&	$38.3\%$ & $-39.5$ \\
	WdW & $\textbf{67.0\%}$&	$41.8\%$&	$52.2\%$ & $-14.8$\\
    WdW-R & $\textbf{69.1\%}$&	$50.0\%$&$50.6\%$ & $-15.6$ \\
\bottomrule
  \end{tabular}
  \caption{Accuracy on various datasets using KV-MemNets (window memory) and GARs}\label{tab:cbt_ga}
  \end{center}
\end{table}




  





  

\begin{table}[t!]
  \begin{center}
  \renewcommand{\arraystretch}{1.2}
  \begin{tabular}{ l c c }
    \hline
    Task & Complete passage & Last sentence \\
    \hline
    CBT-NE&	$22.6\%$&$\textbf{22.8\%}$\\

    CBT-CN&	$\textbf{31.6\%}$&$24.8\%$ \\

    CBT-V&	$\textbf{48.8\%}$&$45.0\%$ \\

    CBT-P&	$34.1\%$&$\textbf{37.9\%}$\\
\bottomrule
  \end{tabular}
  \caption{Accuracy on CBT tasks using KV-MemNets (sentence memory) 
  varying passage size.
  }\label{tab:cbt_sent}
  \end{center}
\end{table}

\begin{table}[ht!]
  \begin{center}
  \renewcommand{\arraystretch}{1.2}
  \begin{tabular}{ c c c c c}
    \hline
    Metric & Full & Q-only & P-only & $\Delta (min)$\\
	\hline
    EM&	$\textbf{70.7\%}$&	$0.6\%$
    & $10.9\%$&$-59.8$ \\    
    F1& $\textbf{79.1\%}$&  $4.0\%$
    &	$14.8\%$&$-64.3$ \\
	\bottomrule
  \end{tabular}
  \caption{Performance of QANet on SQuAD}\label{tab:squad}
  \end{center}
\end{table}

\section{Experimental Results}
\label{sec:results}
\paragraph{bAbI tasks}
Table \ref{tab:tab_babi} shows the results obtained by a Key-Value Memory Network on bAbI tasks by nullifying the information present in either questions or passages. 
On tasks $2,7,13$ and $20$, 
P-only models obtain over $80\%$ accuracy 
with questions randomly assigned.
Moreover, on tasks $3$, $13$, $16$, and $20$,
P-only models match performance 
of those trained on the full dataset.
On task $18$, Q-only models achieve an accuracy of $91\%$,
nearly matching the best performance of $93\%$ achieved by the full model.
These results show that some of bAbI tasks 
are easier than one might think.

\paragraph{Children's Books Test}
On the NE and CN CBT tasks, 
Q-only KV-MemNets obtain an accuracy 
close to the \emph{full} accuracy
and on the Verbs (V) and Prepositions (P) tasks, 
Q-only models outperform the full model (Table \ref{tab:cbt_ga}). 
Q-only Gated attention readers reach accuracy 
of $50.6\%$ and $54\%$ on Named Entities (NE) and Common Nouns (CN) tasks, respectively,
while P-only models reach accuracies of $40.8\%$ and $36.7\%$,
respectively.
%
We note that our models can outperform $16$ 
of the $19$ reported results 
on the NE task in \citet{hill2015goldilocks} 
using Q-only information. 
Table \ref{tab:cbt_sent} shows 
that if we make use of just last sentence 
instead of all $20$ sentences in the passage, 
our sentence memory based KV-MemNet achieve 
comparable or better performance \emph{w.r.t} the \emph{full} model on most subtasks.

\paragraph{CNN}
Table \ref{tab:cbt_ga}, shows the performance 
of Gated Attention Reader on the CNN dataset. 
Q-only and P-only models
obtained $25.6\%$ and $38.3\%$ accuracies respectively,
compared to $77.8\%$ on the true dataset. 
This drop in accuracy could be 
due to the anonymization of entities 
which prevents models from building entity-specific information. 
Notwithstanding the deficiencies noted by \citet{chen2016thorough},
we found that out CNN, 
out all the cloze-style RC datasets that we evaluated, 
appears to be the most carefully designed.

\paragraph{Who-did-What}
P-only models
achieve greater than $50\%$ accuracy 
in both the strict and relaxed setting, 
reaching 
within $15\%$ of the accuracy 
of the \emph{full} model in the strict setting. 
Q-only models also achieve $50\%$ accuracy on the relaxed setting 
while achieving an accuracy of $41.8\%$ 
on the strict setting.  
Our P-only model also outperforms 
all the suppressed baselines 
and $5$ additional baselines 
reported by \citet{onishi2016did}. 
We suspect that the models memorize attributes 
of specific entities, 
justifying the entity-anonymization 
used by \citet{hermann2015teaching} 
to construct the CNN dataset.

\paragraph{SQuAD}
Our results suggest that SQuAD is an unusually carefully-designed and challenging RC task.
The span selection mode of answering requires that models consider the passage
thus the abysmal performance of the Q-only QANet (Table \ref{tab:squad}).
Since SQuAD requires answering by span selection, we construct Q-only variants here by placing answers from all relevant questions 
in random order, filling the gaps with random words.
Moreover, Q-only and P-only models 
achieve F1 scores of only $4\%$ and $14.8\%$ 
resp. (Table \ref{tab:squad}),
significantly lower than $79.1$ on the  
proper task.

\section{Discussion}
\label{sec:discussion}
We briefly discuss our findings, 
offer some guiding principles 
for evaluating new benchmarks and algorithms, 
and speculate on why some of these problems 
may have gone under the radar.
Our goal is not to blame the creators of past datasets 
but instead to support the community 
by offering practical guidance for future researchers.

\paragraph{Provide rigorous RC baselines} 
Published RC datasets should contain reasonable baselines 
that characterize the difficulty of the task, and specifically, 
the extent to which questions and passages are essential.
Moreover, follow-up papers reporting improvements 
ought to report performance both on the full task
and variations omitting questions and passages. 
While many proposed technical innovations
purportedly work by better matching up information in questions and passages, 
absent these baselines 
one cannot tell whether gains come for the claimed reason
or if the models just do a better job of passage classification (disregarding questions).

\paragraph{Test that \emph{full context} is essential} 
Even on tasks where both questions and passages are required, 
problems might appear harder than they really are.
On first glance the the length-$20$ passages
in CBT, might suggest 
that success requires reasoning 
over all $20$ sentences 
to identify the correct answer to each question.
However, it turns out that for some models,
comparable performance can be achieved 
by considering only the last sentence. 
We recommend that researchers provide reasonable ablations
to characterize the amount of context that each model truly requires.

\paragraph{Caution with cloze-style RC datasets} 
We note that cloze-style datasets 
are often created programatically.
Thus it's possible for a dataset 
to be produced, published,
and incorporated into many downstream studies,
all without many person-hours spent manually inspecting the data.
We speculate that, as a result, these datasets tend be subject to less contemplation 
of what's involved in answering these questions 
and are therefore especially susceptible 
to the sorts of overlooked weaknesses described in our study.

\paragraph{A note on publishing incentives}
We express some concern that 
the recommended experimental rigor 
might cut against current publishing incentives.
We speculate that papers introducing datasets 
may be more likely to be accepted at conferences 
by omitting unfavorable ablations than by including them.
Moreover, with reviewers often demanding \emph{architectural novelty},
methods papers may find an easier path to acceptance 
by providing unsubstantiated stories about the \emph{reasons} why a given architecture works than by providing rigorous ablation studies stripping out spurious explanations and unnecessary model components. 
For more general discussions of misaligned incentives 
and empirical rigor in machine learning research,
we point the interested reader to \citet{lipton2018troubling} and \citet{sculley2018winner}.


\bibliography{emnlp2018}

\begin{thebibliography}{33}
\expandafter\ifx\csname natexlab\endcsname\relax\def\natexlab#1{#1}\fi

\bibitem[{Berant et~al.(2014)Berant, Srikumar, Chen, Vander~Linden, Harding,
  Huang, Clark, and Manning}]{berant2014modeling}
Jonathan Berant, Vivek Srikumar, Pei-Chun Chen, Abby Vander~Linden, Brittany
  Harding, Brad Huang, Peter Clark, and Christopher~D Manning. 2014.
\newblock Modeling biological processes for reading comprehension.
\newblock In \emph{Empirical Methods in Natural Language Processing (EMNLP)}.

\bibitem[{Breck et~al.(2001)Breck, Light, Mann, Riloff, Brown, and
  Anand}]{breck2001looking}
Eric Breck, Marc Light, Gideon Mann, Ellen Riloff, Brianne Brown, and Pranav
  Anand. 2001.
\newblock Looking under the hood: Tools for diagnosing your question answering
  engine.
\newblock In \emph{Association for Computational Linguistics (ACL) Workshop on
  Open-Domain Question Answering}.

\bibitem[{Chen et~al.(2016)Chen, Bolton, and Manning}]{chen2016thorough}
Danqi Chen, Jason Bolton, and Christopher~D Manning. 2016.
\newblock A thorough examination of the cnn/daily mail reading comprehension
  task.
\newblock In \emph{Association for Computational Linguistics (ACL)}.

\bibitem[{Clark and Etzioni(2016)}]{clark2016my}
Peter Clark and Oren Etzioni. 2016.
\newblock My computer is an honor student but how intelligent is it?
  standardized tests as a measure of ai.
\newblock \emph{AI Magazine}, 37(1):5--12.

\bibitem[{Dhingra et~al.(2017)Dhingra, Liu, Yang, Cohen, and
  Salakhutdinov}]{dhingra2017gated}
Bhuwan Dhingra, Hanxiao Liu, Zhilin Yang, William Cohen, and Ruslan
  Salakhutdinov. 2017.
\newblock Gated-attention readers for text comprehension.
\newblock In \emph{Association for Computational Linguistics (ACL)}.

\bibitem[{Glockner et~al.(2018)Glockner, Shwartz, and
  Goldberg}]{glockner_acl18}
Max Glockner, Vered Shwartz, and Yoav Goldberg. 2018.
\newblock Breaking nli systems with sentences that require simple lexical
  inferences.
\newblock In \emph{Association for Computational Linguistics (ACL)}.

\bibitem[{Goyal et~al.(2017)Goyal, Khot, Summers-Stay, Batra, and
  Parikh}]{goyal2017making}
Yash Goyal, Tejas Khot, Douglas Summers-Stay, Dhruv Batra, and Devi Parikh.
  2017.
\newblock Making the {V} in {VQA} matter: Elevating the role of image
  understanding in visual question answering.
\newblock In \emph{Computer Vision and Pattern Recognition (CVPR)}.

\bibitem[{Gururangan et~al.(2018)Gururangan, Swayamdipta, Levy, Schwartz,
  Bowman, and Smith}]{N18-2017}
Suchin Gururangan, Swabha Swayamdipta, Omer Levy, Roy Schwartz, Samuel Bowman,
  and Noah~A. Smith. 2018.
\newblock Annotation artifacts in natural language inference data.
\newblock In \emph{North American Chapter of the Association for Computational
  Linguistics (NAACL): Human Language Technologies}.

\bibitem[{Hermann et~al.(2015)Hermann, Kocisky, Grefenstette, Espeholt, Kay,
  Suleyman, and Blunsom}]{hermann2015teaching}
Karl~Moritz Hermann, Tomas Kocisky, Edward Grefenstette, Lasse Espeholt, Will
  Kay, Mustafa Suleyman, and Phil Blunsom. 2015.
\newblock Teaching machines to read and comprehend.
\newblock In \emph{Advances in Neural Information Processing Systems (NIPS)}.

\bibitem[{Hill et~al.(2016)Hill, Bordes, Chopra, and
  Weston}]{hill2015goldilocks}
Felix Hill, Antoine Bordes, Sumit Chopra, and Jason Weston. 2016.
\newblock The {G}oldilocks principle: Reading children's books with explicit
  memory representations.
\newblock In \emph{International Conference on Learning Representations
  (ICLR)}.

\bibitem[{Hirschman et~al.(1999)Hirschman, Light, Breck, and
  Burger}]{hirschman1999deep}
Lynette Hirschman, Marc Light, Eric Breck, and John~D Burger. 1999.
\newblock Deep read: A reading comprehension system.
\newblock In \emph{Association for Computational Linguistics on Computational
  Linguistics (ACL)}.

\bibitem[{Joshi et~al.(2017)Joshi, Choi, Weld, and
  Zettlemoyer}]{joshi2017triviaqa}
Mandar Joshi, Eunsol Choi, Daniel Weld, and Luke Zettlemoyer. 2017.
\newblock Triviaqa: A large scale distantly supervised challenge dataset for
  reading comprehension.
\newblock In \emph{Association for Computational Linguistics (ACL)}.

\bibitem[{Kadlec et~al.(2016)Kadlec, Schmid, Bajgar, and
  Kleindienst}]{kadlec2016text}
Rudolf Kadlec, Martin Schmid, Ond{\v{r}}ej Bajgar, and Jan Kleindienst. 2016.
\newblock Text understanding with the attention sum reader network.
\newblock In \emph{Association for Computational Linguistics (ACL)}.

\bibitem[{Lai et~al.(2017)Lai, Xie, Liu, Yang, and Hovy}]{lai2017race}
Guokun Lai, Qizhe Xie, Hanxiao Liu, Yiming Yang, and Eduard Hovy. 2017.
\newblock Race: Large-scale reading comprehension dataset from examinations.
\newblock In \emph{Empirical Methods in Natural Language Processing (EMNLP)}.

\bibitem[{Lee et~al.(2016)Lee, He, Yih, Gao, Deng, and
  Smolensky}]{lee2015reasoning}
Moontae Lee, Xiaodong He, Wen{-}tau Yih, Jianfeng Gao, Li~Deng, and Paul
  Smolensky. 2016.
\newblock Reasoning in vector space: An exploratory study of question
  answering.
\newblock In \emph{International Conference on Learning Representations
  (ICLR)}.

\bibitem[{Lipton and Steinhardt(2018)}]{lipton2018troubling}
Zachary~C Lipton and Jacob Steinhardt. 2018.
\newblock Troubling trends in machine learning scholarship.
\newblock In \emph{International Conference on Machine Learning (ICML) Machine
  Learning Debates Workshop}.

\bibitem[{Miller et~al.(2016)Miller, Fisch, Dodge, Karimi, Bordes, and
  Weston}]{miller2016key}
Alexander Miller, Adam Fisch, Jesse Dodge, Amir-Hossein Karimi, Antoine Bordes,
  and Jason Weston. 2016.
\newblock Key-value memory networks for directly reading documents.
\newblock In \emph{Proceedings of the 2016 Conference on Empirical Methods in
  Natural Language Processing (EMNLP)}.

\bibitem[{Nguyen et~al.(2016)Nguyen, Rosenberg, Song, Gao, Tiwary, Majumder,
  and Deng}]{nguyen2016ms}
Tri Nguyen, Mir Rosenberg, Xia Song, Jianfeng Gao, Saurabh Tiwary, Rangan
  Majumder, and Li~Deng. 2016.
\newblock {MS MARCO}: A human generated machine reading comprehension dataset.
\newblock \emph{arXiv preprint arXiv:1611.09268}.

\bibitem[{Onishi et~al.(2016)Onishi, Wang, Bansal, Gimpel, and
  McAllester}]{onishi2016did}
Takeshi Onishi, Hai Wang, Mohit Bansal, Kevin Gimpel, and David McAllester.
  2016.
\newblock Who did what: A large-scale person-centered cloze dataset.
\newblock In \emph{Empirical Methods in Natural Language Processing (EMNLP)}.

\bibitem[{Paperno et~al.(2016)Paperno, Kruszewski, Lazaridou, Pham, Bernardi,
  Pezzelle, Baroni, Boleda, and Fernandez}]{paperno2016lambada}
Denis Paperno, Germ{\'a}n Kruszewski, Angeliki Lazaridou, Ngoc~Quan Pham,
  Raffaella Bernardi, Sandro Pezzelle, Marco Baroni, Gemma Boleda, and Raquel
  Fernandez. 2016.
\newblock The lambada dataset: Word prediction requiring a broad discourse
  context.
\newblock In \emph{Association for Computational Linguistics (ACL)}.

\bibitem[{Pe\~nas et~al.(2012)Pe\~nas, Hovy, Forner, Rodrigo, Sutcliffe,
  Sporleder, Forascu, Benajiba, and Osenova}]{penasoverview12}
Anselmo Pe\~nas, Eduard Hovy, Pamela Forner, \'Alvaro Rodrigo, Richard
  Sutcliffe, Caroline Sporleder, Corina Forascu, Yassine Benajiba, and Petya
  Osenova. 2012.
\newblock Overview of qa4mre at clef 2012: Question answering for machine
  reading evaluation.

\bibitem[{Pe{\~n}as et~al.(2011)Pe{\~n}as, Hovy, Forner, Rodrigo, Sutcliffe,
  Forascu, and Sporleder}]{penasoverview11}
Anselmo Pe{\~n}as, Eduard Hovy, Pamela Forner, {\'A}lvaro Rodrigo, Richard
  Sutcliffe, Corina Forascu, and Caroline Sporleder. 2011.
\newblock Overview of qa4mre at clef 2011: Question answering for machine
  reading evaluation.

\bibitem[{Poliak et~al.(2018)Poliak, Naradowsky, Haldar, Rudinger, and
  Van~Durme}]{poliak_nli}
A.~Poliak, J.~Naradowsky, A.~Haldar, R.~Rudinger, and B.~Van~Durme. 2018.
\newblock Hypothesis only baselines in natural language inference.
\newblock In \emph{Joint Conference on Lexical and Computational Semantics
  (*SEM)}.

\bibitem[{Rajpurkar et~al.(2016)Rajpurkar, Zhang, Lopyrev, and
  Liang}]{rajpurkar2016squad}
Pranav Rajpurkar, Jian Zhang, Konstantin Lopyrev, and Percy Liang. 2016.
\newblock Squad: 100,000+ questions for machine comprehension of text.
\newblock In \emph{Empirical Methods in Natural Language Processing (EMNLP)}.

\bibitem[{Richardson et~al.(2013)Richardson, Burges, and
  Renshaw}]{richardson2013mctest}
Matthew Richardson, Christopher~JC Burges, and Erin Renshaw. 2013.
\newblock Mctest: A challenge dataset for the open-domain machine comprehension
  of text.
\newblock In \emph{Empirical Methods in Natural Language Processing (EMNLP)}.

\bibitem[{Sculley et~al.(2018)Sculley, Snoek, Wiltschko, and
  Rahimi}]{sculley2018winner}
D~Sculley, Jasper Snoek, Alex Wiltschko, and Ali Rahimi. 2018.
\newblock Winner's curse? on pace, progress, and empirical rigor.
\newblock In \emph{International Conference on Learning Representations (ICLR)
  Workshop Track}.

\bibitem[{Sukhbaatar et~al.(2015)Sukhbaatar, Weston, Fergus
  et~al.}]{sukhbaatar2015end}
Sainbayar Sukhbaatar, Jason Weston, Rob Fergus, et~al. 2015.
\newblock End-to-end memory networks.
\newblock In \emph{Advances in neural information processing systems (NIPS)}.

\bibitem[{Sutcliffe et~al.(2013)Sutcliffe, Pe{\~n}as, Hovy, Forner, Rodrigo,
  Forascu, Benajiba, and Osenova}]{sutcliffeoverview}
Richard Sutcliffe, Anselmo Pe{\~n}as, Eduard Hovy, Pamela Forner, {\'A}lvaro
  Rodrigo, Corina Forascu, Yassine Benajiba, and Petya Osenova. 2013.
\newblock Overview of qa4mre main task at clef 2013.

\bibitem[{Trischler et~al.(2017)Trischler, Wang, Yuan, Harris, Sordoni,
  Bachman, and Suleman}]{trischler2017newsqa}
Adam Trischler, Tong Wang, Xingdi Yuan, Justin Harris, Alessandro Sordoni,
  Philip Bachman, and Kaheer Suleman. 2017.
\newblock Newsqa: A machine comprehension dataset.
\newblock In \emph{Workshop on Representation Learning for NLP (RepL4NLP)}.

\bibitem[{Wang et~al.(2007)Wang, Smith, and Mitamura}]{wang2007jeopardy}
Mengqiu Wang, Noah~A Smith, and Teruko Mitamura. 2007.
\newblock What is the jeopardy model? a quasi-synchronous grammar for qa.
\newblock In \emph{Empirical Methods in Natural Language Processing and
  Computational Natural Language Learning (EMNLP-CoNLL)}.

\bibitem[{Weston et~al.(2016)Weston, Bordes, Chopra, Rush, van Merri{\"e}nboer,
  Joulin, and Mikolov}]{babi}
Jason Weston, Antoine Bordes, Sumit Chopra, Alexander~M Rush, Bart van
  Merri{\"e}nboer, Armand Joulin, and Tomas Mikolov. 2016.
\newblock Towards ai-complete question answering: A set of prerequisite toy
  tasks.
\newblock In \emph{International Conference on Learning Representations
  (ICLR)}.

\bibitem[{Yang et~al.(2015)Yang, Yih, and Meek}]{yang2015wikiqa}
Yi~Yang, Wen-tau Yih, and Christopher Meek. 2015.
\newblock Wikiqa: A challenge dataset for open-domain question answering.
\newblock In \emph{Empirical Methods in Natural Language Processing (EMNLP)}.

\bibitem[{Yu et~al.(2018)Yu, Dohan, Luong, Zhao, Chen, Norouzi, and
  Le}]{Yu2018QANetCL}
Adams~Wei Yu, David Dohan, Minh-Thang Luong, Rui Zhao, Kai Chen, Mohammad
  Norouzi, and Quoc~V. Le. 2018.
\newblock Qanet: Combining local convolution with global self-attention for
  reading comprehension.
\newblock In \emph{International Conference on Learning Representations
  (ICLR)}.

\end{thebibliography}
\bibliographystyle{acl_natbib_nourl}
\appendix
\end{document}